\documentclass[letterpaper]{article} 
\usepackage{aaai25}  
\usepackage{times}  
\usepackage{helvet}  
\usepackage{courier}  
\usepackage[hyphens]{url}  
\usepackage{graphicx} 
\usepackage{natbib}  
\usepackage{caption} 
\usepackage{algorithm}
\usepackage{algorithmicx}
\usepackage[noend]{algpseudocode}
\usepackage{subcaption}
\usepackage{pifont}
\usepackage{amsmath}
\graphicspath{{figures/}}
\usepackage{multirow}
\usepackage{framed}
\usepackage{tikz}
\usepackage[framemethod=TikZ]{mdframed}
\usepackage{enumitem}
\usepackage{xcolor}
\graphicspath{{figures/}}

\usepackage{xspace}
\newcommand{\name}{\texttt{$\mu$MOEA}\xspace}
\usepackage{newfloat}
\usepackage{listings}
\DeclareCaptionStyle{ruled}{labelfont=normalfont,labelsep=colon,strut=off} 
\floatstyle{ruled}
\newfloat{listing}{tb}{lst}{}
\floatname{listing}{Listing}
\title{An LLM-Empowered Adaptive Evolutionary Algorithm For Multi-Component Deep Learning Systems}
\author{
    Haoxiang Tian\textsuperscript{\rm 1,2}\thanks{This work was done at NTU as a visiting student.},
    Xingshuo Han\thanks{Xingshuo Han and Guoquan Wu are Corresponding authors.}\textsuperscript{\rm 3},
    Guoquan Wu\footnotemark[2]\textsuperscript{\rm 1,4},
    An Guo\textsuperscript{\rm 5},
    Yuan Zhou\textsuperscript{\rm 6},
    Jie Zhang\textsuperscript{\rm 7}, 
    Shuo Li\textsuperscript{\rm 1},
    Jun Wei\textsuperscript{\rm 1,4},
    Tianwei Zhang\textsuperscript{\rm 2}
}
\affiliations{
    \textsuperscript{\rm 1}Key Lab of System Software at CAS\thanks{CAS is the abbreviation of Chinese Academy of Sciences}, State Key Lab of Computer Science at ISCAS\thanks{ISCAS is the abbreviation of Institute of Software at CAS}, University of CAS, Beijing\\
    \textsuperscript{\rm 2} Nanyang Technological University, Singapore \\
    \textsuperscript{\rm 3} Continental-NTU Corporate Lab, Singapore\\
    \textsuperscript{\rm 4}Nanjing Institute of Software Technology, University of CAS, Nanjing\\ 
    \textsuperscript{\rm 5}Nanjing University, Nanjing \\
    \textsuperscript{\rm 6}Zhejiang Sci-Tech University, Hangzhou \\ 
    \textsuperscript{\rm 7}CFAR and IHPC, A*STAR, Singapore\\
    \{tianhaoxiang20, gqwu, lishuo19, wj\}@otcaix.iscas.ac.cn, xingshuo001@e.ntu.edu.sg, guoan218@smail.nju.edu.cn, yuanzhou@zstu.edu.cn, zhang\_jie@cfar.a-star.edu.sg, tianwei.zhang@ntu.edu.sg
}

\usepackage{bibentry}

\begin{document}

\maketitle

\begin{abstract}
Multi-objective evolutionary algorithms (MOEAs) are widely used for searching optimal solutions in complex multi-component applications. Traditional MOEAs for multi-component deep learning (MCDL) systems face challenges in enhancing the search efficiency while maintaining the diversity. To combat these, this paper proposes the first LLM-empowered adaptive evolutionary search algorithm to detect safety violations in MCDL systems. Inspired by the context-understanding ability of Large Language Models (LLMs), our approach promotes the LLM to comprehend the optimization problem and generate an initial population tailed to evolutionary objectives. Subsequently, it employs adaptive selection and variation to iteratively produce offspring, balancing the evolutionary efficiency and diversity. During the evolutionary process, to navigate away from the local optima, our approach integrates the evolutionary experience back into the LLM. This utilization harnesses the LLM's quantitative reasoning prowess to generate differential seeds, breaking away from current optimal solutions. We evaluate our approach in finding safety violations of MCDL systems, and compare its performance with state-of-the-art MOEA methods. Experimental results show that our approach can significantly improve the efficiency and diversity of the evolutionary search.
\end{abstract}

%

\section{Introduction}

Multi-component deep learning systems (MCDL systems) are intricate and characterized by significant uncertainty due to their complexity. These systems often involve multiple interacting modules, each with its own set of parameters and behaviors, leading to unpredictable emergent properties. In real-world scenarios, MCDL systems are increasingly being deployed in domains with substantial societal impact, such as autonomous vehicles, healthcare, and financial services. For instance, autonomous driving systems integrate various components (e.g., object detection, path planning, and decision-making), where even minor faults in one component can lead to catastrophic outcomes \cite{bojarski2016end}. Therefore, it is crucial to detect as many safety violations as possible in these systems to mitigate risks and ensure reliability \cite{goodfellow2017attacking}.

Multi-objective evolutionary algorithms (MOEAs) are widely applied to search for elite solutions to find safety violations in MCDL systems \cite{tian2022mosat, abdukhamidov2023unveiling, abdukhamidov2023microbial}. They can be formulated as multi-objective optimization problems \cite{zhou2011multiobjective}. In practical evolutionary search solutions based on genetic algorithms, these multiple objectives correspond to various perspectives (e.g., maximize the quality of the solution, improve the diversity of solutions, balance the cost of solutions against benefits) \cite{ehrgott2014minmax, long2014constraint}. However, there are trade-offs between these objectives as they conflicts with each other in many real-world scenarios.

The evolutionary search process generally consists of three steps: 1) initializing the initial population, 2) evaluating each generated individual with a defined fitness function, and 3) selecting high-fitness individuals to conduct variation operators to generate offspring iteratively. However, existing MOEAs for detecting safety violations in MCDL systems~\cite{tian2022mosat} face two challenges that have not been well addressed.

\begin{itemize}
\item \textbf{Challenge-1}: The initialization of the population highly affects the search efficiency for elitist solutions.
However, in many MOEAs, the initial population is created by random initialization of parameters in the entire search space, which is highly contingent and uncertain.

\item \textbf{Challenge-2}: The evolutionary search process is prone to get stuck at local optima. In existing MOEAs, the individuals of each generation are generated by the high-fitness individuals retained from previous generations, which tend to cause convergence prematurely and result in a large number of iterations that only find few similar safety violations of the MCDL systems. 
\end{itemize}

The goal of this paper is to overcome the above challenges and improve the evolutionary search efficiency and diversity. Large Language Models (LLMs), such as GPT-4 from OpenAI \cite{roumeliotis2023chatgpt}), have demonstrated remarkable abilities
in language understanding and quantitative reasoning \cite{romera2024mathematical, zhang2022opt}. So we propose to leverage these capabilities to generate the initial population and navigate the evolutionary search process away from the local optima.
However, despite the LLMs' expertise in interacting with humans,
it is infeasible to directly apply them to search for optimal and diverse solutions for detecting safety violations of MCDL systems. This is because MCDL systems normally have high dimensionality and complexity, making it difficult for LLMs to fully and accurately understand the search space. Additionally, since LLMs may not have enough specific domain knowledge about the MCDL systems, they will arbitrarily modify existing solutions and make up unreasonable solutions, rendering them less effective. 

We design \name, the first LLM-empowered adaptive evolutionary methodology. \name uses LLM's ability in language understanding to better comprehend the search task, which creates the individuals of the initial population considering the objectives instead of random initialization, \textbf{thus addressing C1}. Based on the initial population, to balance the search efficiency and diversity, \name introduces adaptive selection and a suite of adaptive variations, which can dynamically adjust the mutation and crossover probabilities based on the feedback from the search process and the scores of chromosomes of individuals on different objectives.
During the search process, when it gets stuck, \name feeds back the evolutionary experience into the LLM, harnessing its quantitative reasoning ability to generate differential seeds to break out of local optimal solutions, \textbf{thus addressing C2}. We evaluate the effectiveness of \name in the task of searching for solutions to detect safety violations in MCDL systems (represented by the industrial autonomous driving system), and compare it with the state-of-the-art (SOTA) method based on multi-objective genetic algorithm (NSGA-II). Experimental results show that \name can find more diverse elitist solutions more efficiently.

\section{Background}
\subsection{Multi-Component Deep Learning Systems}
MCDL systems are characterized by their intricate internal logic, extensive interactions, and high coupling among various components \cite{amodei2016concrete, varshney2016engineering}. Their complexity and opaqueness are further exacerbated by the substantial uncertainty, high degree of interdependence, and unpredictable nature of interactions across different deep learning models within the systems. Thus an MCDL system is often referred to as a “black box” \cite{hassija2024interpreting}, making it challenging to thoroughly detect potential safety issues under varying conditions. It is necessary and urgent to have effective methods for the examination and detection of the internal problems in MCDL systems, without requiring a detailed understanding of the intricate workings.
 
A proven strategy to detect safety violations in MCDL systems is to generate solutions that simulate the diverse operational conditions and assess the system's behaviors (including responses, decisions, operations/actions) \cite{tian2022generating} to validate whether it adheres to the safety specifications \cite{borg2018safely, guo2024sovar}. Ensuring the safety and reliability of MCDL systems requires diverse solutions to detect various potential vulnerabilities and failures of MCDL systems in a wide range of conditions \cite{asharf2020review, tian2022mosat}. However, given the large state space of MCDL systems, traditional methods struggle to cover more possible cases efficiently. Thus, there is a growing need for more adaptive and comprehensive approaches to safety assessment of MCDL systems.

\subsection{MOEAs For MCDL Systems}
MOEAs, represented by NSGA-II \cite{deb2002fast}, are widely used in detecting safety violations in MCDL systems. They are capable of exploring the vast and complex space of system behaviors and identify any misbehaviors \cite{mishra2019test}. This is achieved with evolving and optimizing solutions to cover possible situations \cite{wirsansky2020hands} where the system behaviors violate the safety specifications. 

The overall process of an MOEA (e.g., NSGA-II) is given as follows.
\begin{itemize}
    \item \textbf{Initial population}: NSGA-II commences by initializing a population of $N$ solutions, which is randomly generated within the solution space.
    \item \textbf{Fitness function}: given a solution space $S$ and objective functions $f_1, f_2, ..., f_m$, the multi-objective optimization can be formulated as: 
    $max_{x\in S} \{ f_1(x), f_2(x), ..., f_m(x) \}$.
    To detect the safety violations in MCDL systems, the multiple objectives commonly include maximizing the fault detection and solution diversity.
    \item \textbf{Ranking-based selection}: in the $i$-th generation, the solutions are evaluated by the fitness function and sorted by the crowding distance. $P_i$ consists of the non-dominated solutions obtained by each Pareto frontier. NSGA-II selects $k$ solutions from $P_i$.
    \item \textbf{Variation}: NSGA-II calculates the crossover probability and mutation probability for each selected solution, and compares the probability with the threshold of variation to determine whether to conduct multi-point crossover or value mutation on it. The probability of variation is calculated by a random value in (0, 1) and the threshold is pre-defined by a fixed value in (0, 1).
    \item \textbf{Iterative generations}: after generating $N$ offspring solutions, the next generation’s population is determined by selecting the best $N$ solutions from the current population $P_i$ and the offspring population $P_{i+1}$. NSGA-II iteratively searches and refines candidate solutions based on their performance against these defined objectives.
\end{itemize}

\section{Methodology}

\begin{algorithm}[h]
    \begin{small}
    \caption{LLM-empowered adaptive evolutionary search}
        \label{alg:overall}
    \begin{algorithmic}[1]
        \Ensure	The solution set SCR
        \Require The form of solution AE, starting prompt $pt$
        \State P $\leftarrow \emptyset$, P $\leftarrow$ P $\bigcup$ LLM\_generate(AE, $pt$)
        \While{not TerminationCondition()}
            \State TS, SC, MS $\leftarrow \emptyset$
            \For {$p_i \in$ P[-1]}
                \State execute $p_i$
                \If{$\exists$ ego safety violation in $p_i$}
                    \State SCR $\leftarrow$ SCR $\bigcup p_i$
                \EndIf
            \EndFor
            \State calculate fitness S
            \State TS $\leftarrow$ TS $\bigcup$ S
            \State SC, MS $\leftarrow$ ADAPTIVE SELECTION(P, S)
            \State P $\leftarrow$ P $\bigcup$ ADAPTIVE VARIATION(SC, MS)
            \State prompt = $pt$ + generate\_feedback(P, S)
            \State P $\leftarrow$ P $\bigcup$ LLM\_generate(prompt)
        \EndWhile
        \State return SCR
    \Procedure {Adaptive Selection}{P, S}
        \State $SC, MS \leftarrow \emptyset$
        \State $CR_{r}, MR_{r} = calculate\_variation\_rate(S)$
        \For {$p_i, p_j \in$ P}
            \State select fitness $s_i, s_j \in$ S
            \State {$c_{i,j} = generate\_crossover\_probability(s_i, s_j, S)$}
            \If {$c_{i,j} > CR_r$}
                \State {$SC \leftarrow SC \bigcup (p_i, p_j)$}
            \EndIf
        \EndFor
        \For {$p_i \in$ P}
            \State {$m_i = generate\_mutation\_probability(s_i, S)$}
            \If {$m_i > MR_r$}
                \State {$MS \leftarrow MS \bigcup p_i$}
            \EndIf
        \EndFor
        \State return SC, MR 
    \EndProcedure
    \Procedure {Adaptive Variation}{SC, MS}
        \For {$x_i \in SC$}
            \State {$p_i^\prime,p_j^\prime \leftarrow adaptive\_crossover(x_i)$}
            \State {$PN \leftarrow PN \bigcup \{p_i^\prime,p_j^\prime$\}}
        \EndFor
        \For {$p_i \in MS$}
            \State {$p_i^\prime \leftarrow adaptive\_mutation(p_i)$}
            \State {$PN \leftarrow PN \bigcup p_i^\prime$}
        \EndFor
        \State return $PN$
    \EndProcedure
    \end{algorithmic}
    \end{small}
\end{algorithm}

\begin{figure}[t]
  \centering
  \includegraphics[scale=0.45]{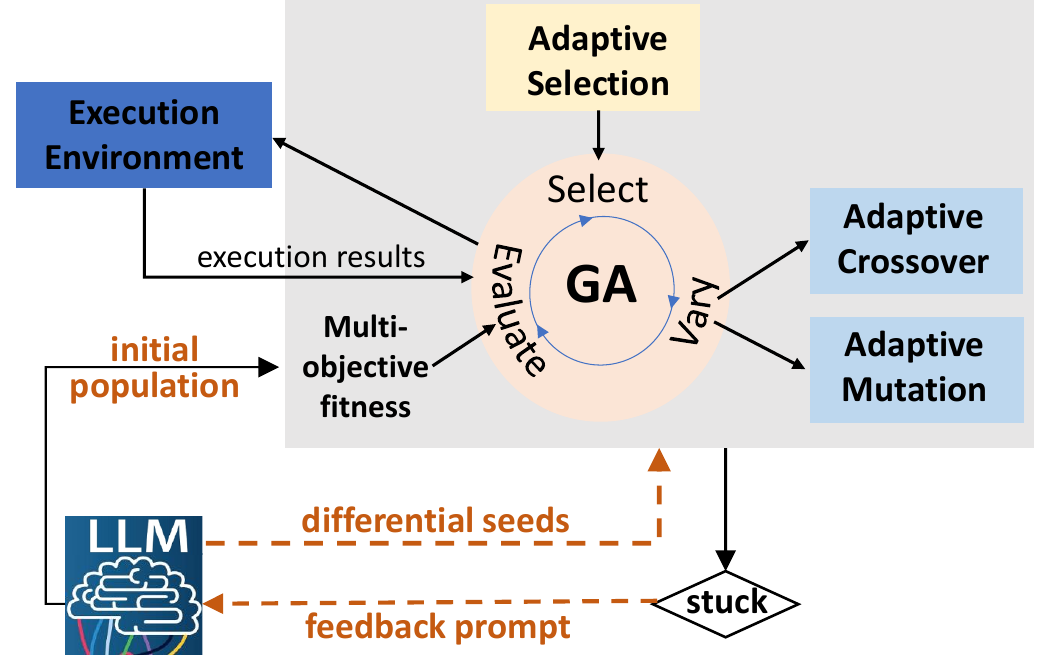}
  \caption{Overall workflow of \name}
  \label{overview}
\end{figure}

We introduce \name, a novel LLM-empowered adaptive evolutionary search algorithm for MCDL systems. The detailed process of \name is illustrated in Algorithm \ref{alg:overall} and Figure~\ref{overview}. It consists of three steps: instructing the LLM to create the initial population (line 2), evolving the population adaptively to search for optimal solutions (line 3-12), guiding the LLM to generate differential seeds based on the feedback of the evolutionary process (line 13-14). Below we give detailed explanation of each step. 

\subsection{Instructing LLM to Create Initial Population}

When starting an evolutionary search, instead of randomly initializing $N$ solutions as the initial population, \name makes the LLM understand the evolutionary task and create the initial population by considering the objectives of the search. This is achieved with the linguistic prompt. An example of the prompt patterns are shown in Table~\ref{first_prompt}, which are designed from the following aspects:
\begin{itemize}
    \item The structure of the solution, including essential blocks of the test case, the participant or element that each block represents. This promotes the LLM to understand the correct representation of solutions.
    \item The keyword/statement, essential attributes and their value ranges of each block. This promotes the LLM to reason within the feasible ranges.
    \item The examples of feasible solutions and their explanations of the requirements. This promotes the LLM to learn the test requirements.
\end{itemize}
These components of the prompt pattern enable $\mu$MOEA to facilitate the LLM to understand the requirement and learn how to generate individuals for the initial population. This also addresses the limitation of the LLM that struggles with high-dimensional search spaces.

\begin{table}[h]
\begin{tabular}{c|c}
\hline
\textbf{Prompt}                                                  & \textbf{Sample of linguistic patterns}                                                                                                                                                                                     \\ \hline
\begin{tabular}[c]{@{}c@{}}Starting\\ Prompt\end{tabular}             & \multicolumn{1}{l}{\begin{tabular}[c]{@{}l@{}}You are an expert of \textless{} MCDL system\textgreater\\ We want you to generate \textless{}N\textgreater solutions\\ for the system\end{tabular}}                   \\ \hline
\begin{tabular}[c]{@{}c@{}}Task Under-\\-standing\\ Prompt\end{tabular} & \begin{tabular}[c]{@{}c@{}}\textless{}The form of the solution\textgreater\\ \textless{}Introduction of parameters of solution\textgreater\\ \textless{}Introduction of the elitism of solution\textgreater{}\end{tabular} \\ \hline
\end{tabular}
\caption{Prompt patterns for LLM-based initial population}
\label{first_prompt}
\end{table}

\name sends the starting prompt along with the task understanding prompt into the LLM, to obtain the initial population. Taking the autonomous driving system as an example, a template of ``introduction of the elitism of solution'' is given as follows: the trajectories of NPC vehicles and pedestrians are required to disturb the ego vehicle's driving path. They need to be different from each other, and the waypoints of them need to involve different lanes.

\subsection{Evolving Population Adaptively for Optimal Solutions}

Based on the initial population, \name adaptively evolves them to search for diverse optimal solutions to detect safety violations in the MCDL system. Each solution is encoded as an individual $P_i = \{ C_1, C_2, ... C_n \}$, where $C_n$ represents the $n$-th chromosome consisting of a series of genes (a chromosome commonly corresponds to an element or object in the solution, and a gene corresponds to an action or operation). Different from existing MOEAs where the parent individuals and their chromosomes undergo the same level of mutation and crossover, \name adopts \textit{adaptive selection} and \textit{adaptive variations}, to improve the heritability of the elite features and search efficiency. 

(1) For each generation, \name builds improved Pareto-optimal solutions considering multiple objectives to measure the potential of solutions to expose safety violations of the MCDL system. The criticality metric is used to evaluate how close the tested MCDL system's behavior is to safety violations. The multi-objectives of $\mu$MOEA include criticality and diversity. Practically, criticality is formulated as: 
\begin{equation}
f_c^{s_i} = \min_{t \in s_i} \{ SV_{MCDL}^t \} \nonumber
\end{equation}
where $SV$ is the distance to safety violations. The diversity metric is used to evaluate the coverage of the test cases generated by previous generations. Practically, diversity is formulated as: 
\begin{equation}
f_d^{s_i} = \frac{\sum_{j=1}^{\lvert PN \lvert} ED_{s_i, s_j}}{\lvert PN \lvert},  ED_{s_i,s_j}=\frac{\sum_{n=1}^{|s_i|} \sum_{m=1}^{|s_j|} TD_{s_i^n, s_j^m}} {|s_i|*|s_j|}
\nonumber
\end{equation}
\begin{equation}
TD_{s_i^n, s_j^m}=\sum_{k=0}^{\alpha} \sqrt{(x_{s_i^n.k}-x_{s_j^m.k})^2+(y_{s_i^n.k}-y_{s_j^m.k})^2}
\nonumber
\end{equation}
The fitness function is represented as: 
\begin{equation}
S \leftarrow arg_{i \in G} \left\{min f_c^{s_i}, max f_d^{s_i} \right\} \nonumber
\end{equation}
where G represents the current generation; $f_c$ is the metric that evaluates how close the tested MCDL system's behavior is to safety violations; $f_d$ is the metric that evaluates the coverage of test cases generated by previous generations.

(2) Based on (1), \name performs \textit{adaptive selection}, which selects high-fitness solutions based on the fitness evaluated by the multiple objectives as parents for variation to generate offspring, iteratively producing Pareto-optimal solutions. To determine the solutions for crossover and mutation, \name varies the probabilities of crossover and mutation adaptively in response to the fitness values of the current population, which promotes high-fitness solutions having larger crossover probabilities and low-fitness solutions having larger mutation probabilities. 

(3) For the selected individuals, \name performs \textit{adaptive variation}: which includes adaptive crossover and adaptive mutation. It dynamically selects different types of variation operations based on the ranking of chromosomes' scores in the population on each objective, which makes elite chromosomes better spread their features into the offspring with more different chromosomes, and makes inferior chromosomes more disrupted. Below we give detailed description of these two steps. 

\subsubsection{Adaptive Selection.}
For the current population and its parents, \name adaptively determines the candidate solutions to conduct crossover and mutation according to their fitness values and fitness level of the population.
For the solution $s_i$ with fitness $f_i$ in the population $p_n$, $f_{max}$ and $f_{min}$ represent the maximal and minimal fitness values in $p_n$ respectively. For each population $p_n$, \name calculates the average values of fitness, represented as $\overline{f}$. The mutation probability of $s_i$ is $PM_i$. For the two solutions $s_i$ and $s_j$, their crossover probability is represented as $PC_{i,j}$, and $f^{'}_{i,j}$ is the larger of the
fitness values.

For the selection of solutions for crossover, the higher the fitness value of one solution, the larger the probability of crossover between it and the other candidate solution. For $s_i$ and $s_j$, The closer $f^{'}_{i,j}$ is to $f_{max}$, the larger the $PC_{i,j}$ is. The crossover probability of $s_i$ and $s_j$ is computed as follows, where $0<k_2, k_4\le 1$. If $PM_i \ge threshold_m^n$, \name will conduct adaptive mutation on it. 

\begin{equation}
PM_c = \min \{k_2 (f_i-f_{min})/(f_{max}-\overline{f}), k_4\} \nonumber
\end{equation}

For the selection of solutions for mutation, the smaller the fitness value of the solution, the higher the probability of mutating its parameters. For $s_i$, the closer $f_i$ is to $f_{min}$, the larger $PM_i$ is. The mutation probability of $s_i$ is computed as follows, where $0<k_2, k_4\le 1$. If $PM_i \ge threshold_m^n$, \name conduct adaptive mutation on it.
\begin{equation}
    PM_i = \min \{k_2 (f_{max}-f_i)/(\overline{f}-f_{min}), k_4\}, \nonumber
\end{equation}

\begin{table*}[t]
\renewcommand{\arraystretch}{1.3}
\centering
\begin{tabular}{c|l}
\hline
\textbf{Rule} & \multicolumn{1}{c}{\textbf{Sample of feedback prompt}}                                                                                                                                                                                                                                                                                                                                                                                  \\ \hline
1             & \begin{tabular}[c]{@{}l@{}}Each solution in \textless{}$SE$\textgreater exposed a safety violation of \textless{}MCDL system\textgreater{}. So they are what we want.\\ No safety violation occurred in \textless{}$SN$\textgreater{}, which are not required by us. We want you to\\ generate \textless{}$N$\textgreater solutions that can expose safety violations and differentiate from \textless{}$SE$\textgreater{}\end{tabular} \\ \hline
2             & \begin{tabular}[c]{@{}l@{}}The solutions in \textless{}$R$\textgreater are not different enough from \textless{}$SE$\textgreater{}. Please re-generate to create new\\ solutions that have high potential to expose safety violations of \textless{}MCDL system\textgreater{}.\end{tabular}                                                                                                                                             \\ \hline
\end{tabular}
\caption{Rules for feedback prompt generation}
\label{feedback_prompt}
\end{table*}

To disrupt the solutions with above-average fitness values to search the spaces for the region with global optimum, and ensure that all solutions with subaverage fitness values compulsorily undergo mutation, we use a value of 0.6 for $k_1$ and $k_2$, and 1.0 for $k_3$ and $k_4$. These values can be changed according to the actual needs. 

Since the threshold of variation ($threshold_n$) has great effects on the overall variation of the population $p_n$, different from the MOEAs that pre-define a fixed value for the threshold, \name computes $threshold_n$ for each population $p_n$ adaptively based on the population's overall level of fitness values. When the fitness values of the solutions in $p_n$ converge, \name decreases the $threshold_n$ to facilitate the crossover and mutation to create more different offspring. Similarly, if the fitness values of the population scatter, \name increases the $threshold_n$ to accelerate the convergency to find an optimal solution. The calculation of $threshold_n$ is given as follows, where $0<c_1,m_1\le 1$.
\begin{equation}
threshold_n = c_1(f_{max}-\overline{f}) + m_1(\overline{f}-f_{min}) \nonumber
\end{equation}

\subsubsection{Adaptive Variation.}
\label{variation}
This includes adaptive crossover and adaptive mutation. The variation strategy has a higher probability that the generated offspring can integrate the advantages of parents in convergence and diversity.

For adaptive crossover, given two candidate solutions of crossover, based on the objective that the solution ranks highest in the population, the chromosome with the highest value on the objective is selected for crossover using the single-point crossover with a random chromosome in the other candidate solution.

For adaptive mutation,  given the candidate solution of mutation, different types of mutation operations are dynamically determined based on the fitness values of chromosomes in the solution. If the chromosome has a high score on any objective in the population, \name adjusts its parameters slightly (e.g., modifying the parameters of some genes on it) to better explore the surrounding space. Otherwise, \name makes major changes to it, e.g., changing the combinations or sequences of genes on it, adding new actions/operations into it, replacing some genes with new actions/operations.

\subsection{Guiding LLM to Generate Differential Seeds}

For the generated solutions, \name runs them to detect the safety violations in the MCDL system. During the adaptive evolutionary search, we find that as the iterations increase, the evolutionary search is prone to falling into local optimality, causing the newly generated solutions similar to those optimal solutions generated by previous generations. To solve the issue, when the evolutionary search gets stuck, \name generates differential seed solutions for the next generation, which encourages the exploration of more diverse solutions. 

Specifically, when the high-fitness solutions in $t$ consecutive generations remain the same, \name selects the optimal solutions generated by the previous iterations (collecting their chromosomes in $SE$), and generates the feedback prompt using Rule 1 in Table~\ref{feedback_prompt} (where $SN$ represents the chromosomes of non-optimal solutions). The prompt pattern of Rule 1 is to promote the LLM to learn the characteristics of previous evolutionary iterations, and then create differential seed solutions leveraging its reasoning capability. 

Considering that the LLM is typically accustomed to generating outputs that resemble the examples provided in the input, \name examines the differences between the LLM's generated solutions to the input solutions. For the solution $s_i$, its difference from the input solutions is calculated as: $d_{i}=(\sum_{x_o \in (SN \bigcup SE)} ED_{s_i, x_o})/r$, where $ED$ represents the Euclidean Distances between two solutions. For the LLM's generated solutions that do not meet the difference requirements with the previous solutions, \name generates the prompt based on Rule 2 in Table~\ref{feedback_prompt} to make the LLM re-generate qualified differential seed solutions. 

\section{Evaluation}
To evaluate the effectiveness and advancement of \name, we apply it to search for solutions that detect safety violations of the representative MCDL system, and compare \name's performance to the SOTA method.

\subsection{Experiment Setup}
\textbf{MCDL System.}
Autonomous driving systems (ADSs) exemplify a prototypical case of multi-component deep learning (MCDL) systems, comprising various components built upon multiple deep learning models. These components and models engage in high-frequency communication and input-output interactions. Given the considerable social implications of autonomous driving technology, detecting safety violations of ADSs is of substantial importance.

We select the industrial full-stack ADS, Baidu Apollo \cite{apolloauto} to evaluate the ability of \name in finding safety violations of MCDL systems, due to the representativeness, practicality and advancedness. (1) Representativeness. Apollo ranks among the top 4 leading industrial ADS developers \cite{rank} (the other three ADSs, Waymo, Ford, and Cruise, are not released publicly). (2) Practicality. Apollo can be readily installed on vehicles for driving on public roads \cite{launch} (it has provided self-driving services for real vehicles \cite{selfhighway, baidutaxi}). (3) Advancedness. Apollo is actively and rapidly updated (the releases of Apollo update on a weekly basis).

\noindent\textbf{Test Platform.}
We conducted the experiments on Ubuntu 20.04 with 500 GB memory, an Intel Core i7 CPU, and an NVIDIA GTX2080 TI. SORA-SVL \cite{sora-svl} (an end-to-end AV simulation platform which supports connection with Apollo) and San Francisco map are selected to execute the generated solutions. 
During the experiments, all modules of Apollo are turned on, including perception module, localization module, prediction module, routing module, planning module, and control module.

\noindent\textbf{Evaluation Metrics.}
To evaluate the effectiveness of the method in detecting diverse safety violations of the ADS, the metrics include the following aspects:

\begin{itemize}
    \item How many types of safety violations are detected?
    \item How many solutions are generated on average to detect one safety violation?
    \item How long does it take on average to detect the first safety violation and all found types of safety violations?
\end{itemize}



\subsection{Effectivenss of \name}

We run \name for 24 hours to detect safety violations of Apollo. For the found safety violations, we analyze their root causes by locating the incorrect operations of modules in Apollo. Furthermore, based on the analysis, we classify the found safety violations into distinct types. To account for the randomness, the experiments are repeated five times and the average results are provided as follows.

For each run, on average, 3756 solutions (min 3346 and max 4015) are generated by \name and 313 (min 292 and max 355) out of them have safety violations of Apollo. \name can detect 10 distinct types of safety violations of Apollo, which are all revealed in the first 14 hours.

\begin{figure}[t]
  \centering
  \includegraphics[scale=0.7]{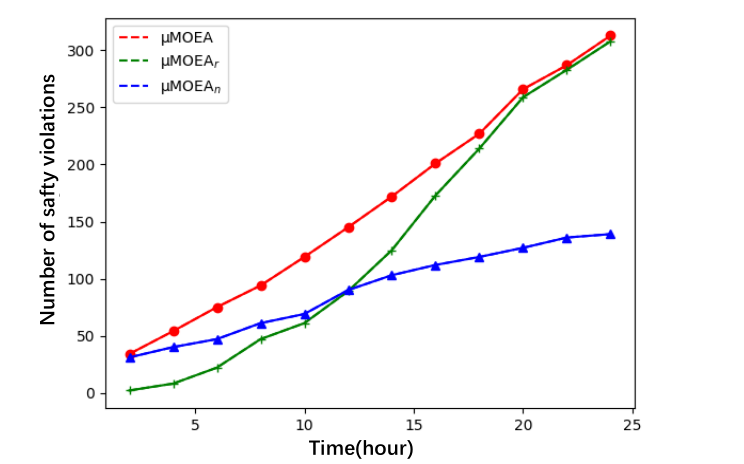}
  \caption{The number of found safety violations over time}
  \label{running}
\end{figure}

To evaluate the benefit of the LLM-based initial population creation and differential seed generation, we conduct the ablation experiments of \name. Two variant versions $\mu MOEA_r$ and $\mu MOEA_n$ are implemented. $\mu MOEA_r$ creates the initial population by random initialization of solutions, and $\mu MOEA_n$ evolves the solutions without differential seeds. We run \name, $\mu MOEA_r$ and $\mu MOEA_n$ for the same amount of time, and compare their effectiveness and efficiency. The results are shown as Table~\ref{abolation} (where SV is the abbreviation for safety violation) and Figure~\ref{running}.

\begin{table}[t]
\begin{tabular}{c|c|c|c}
\hline
\multicolumn{1}{l|}{}                                                  & \textbf{\name} & $\mu$MOEA$_r$ & \textbf{$\mu$MOEA$_n$} \\ \hline
\begin{tabular}[c]{@{}c@{}}types of\\detected SV\end{tabular}            & 10             & 10                 & 6                  \\ \hline
\begin{tabular}[c]{@{}c@{}}number of solution\\to detect one SV\end{tabular}            & 12             & 12.7                 & 24.3                  \\ \hline
\begin{tabular}[c]{@{}c@{}}time to detect\\ the first SV\end{tabular}    & 11min          & 39min              & 17min              \\ \hline
\begin{tabular}[c]{@{}c@{}}time to detect\\all types of SVs\end{tabular} & 14h            & 15h                & 22h                \\ \hline
\end{tabular}
\caption{Results of \name and ablation baselines}
\label{abolation}
\end{table} 


$\mu MOEA_r$ can detect 10 types of safety violations of Apollo, and $\mu MOEA_n$ can detect 6 types of safety violations of Apollo. On average, in the 24-hour run, $\mu MOEA_r$ generates 3922 solutions (min 3853 and max 4014), and 308 of them (min 289 and max 315) detect safety violations of Apollo. For $\mu MOEA_n$, it generates 3352 solutions (min 3099 and max 3480), and 138 of them (min 121 and max 150) detect safety violations of Apollo. 

Table~\ref{abolation} shows that $\mu MOEA_r$ takes the most time to detect the first safety violation of Apollo. It can be seen from Figure~\ref{running}(a) that in early-generation solutions, the number of safety violations detected by $\mu MOEA_r$ is the least. We can conclude that \name's creation of initial population creation helps generate better initial population than random initialization. 

For $\mu MOEA_n$, Table~\ref{abolation} shows that the number of safety violation types detected by it is fewer than \name and $\mu MOEA_r$, and it takes more time to detect all found types of safety violations. The average Euclidean Distance across the 
detected different safety violation types of \name is 81.10 meters, and that of $\mu MOEA_r$ is 69.72 meters. From Figure~\ref{running}(b), we can see that as the iterations of evolutionary search increase, the growth of safety violations detected by $\mu MOEA_n$ is slowest. We can conclude that the differential seeds of \name can help solve the local optimal and detect more diverse types of safety violations.

It's worth noting that, during the iterations of \name and $\mu MOEA_r$, after a few types of safety violations have been found, the detection of safety violations grows faster. We analyze that it benefits from the feedback-based differential seed generation. As the evolutionary iterations increase, the solutions of feedback increase. \name can better learn more experience about the characteristics of optimal solutions, which can improve the quality of the generated differential seed solutions.

\subsection{Advancement of \name}
We evaluate \name in comparison to the SOTA method that uses the multi-objective genetic algorithm (NSGA-II) to detect safety violations of Apollo: MOSAT \cite{tian2022mosat}. MOSAT generates the first population by randomly initialized individuals. Based on them, MOSAT uses multi-objective genetic algorithm to
csearch for optimal and diverse solutions. The individuals are evaluated by multi-objective fitness function, which contains the elitism and diversity. MOSAT determines crossover probability and mutation probability of parent individuals by random rates and the fixed variation threshold. The variation operators that are defined to manipulate individuals include uniform crossover and mutation.

We run MOSAT on the same road in San Francisco as \name. For the sake of fairness, in each 24-hour running, the number of individuals in each generation of MOSAT and \name are the same. The comparison results of \name and MOSAT are shown as Table~\ref{comparison}.

\begin{table}[h]
\centering
\begin{tabular}{|cl|c|c|}
\hline
\multicolumn{2}{|c|}{Approach}                                                                                                        & \textbf{\name} & \textbf{MOSAT} \\ \hline
\multicolumn{2}{|c|}{types of detected SV}                                                                                            & 10                            & 6              \\ \hline
\multicolumn{1}{|c|}{\multirow{3}{*}{\begin{tabular}[c]{@{}c@{}}number of solutions\\ to detect one SV\end{tabular}}}             & min & 10.1                          & 61.0           \\ \cline{2-4} 
\multicolumn{1}{|c|}{}                                                                                                          & max & 13.7                          & 65.9           \\ \cline{2-4} 
\multicolumn{1}{|c|}{}                                                                                                          & avg & 12                            & 62.1           \\ \hline
\multicolumn{1}{|c|}{\multirow{3}{*}{\begin{tabular}[c]{@{}c@{}}time to detect the\\ first SV (min)\end{tabular}}}                & min & 1                             & 2              \\ \cline{2-4} 
\multicolumn{1}{|c|}{}                                                                                                          & max & 12                            & 28             \\ \cline{2-4} 
\multicolumn{1}{|c|}{}                                                                                                          & avg & 7                             & 16             \\ \hline
\multicolumn{1}{|c|}{\multirow{3}{*}{\begin{tabular}[c]{@{}c@{}}time to detect all\\ found types of SVs\\ (hour)\end{tabular}}} & min & 11.3                          & 16.0           \\ \cline{2-4} 
\multicolumn{1}{|c|}{}                                                                                                          & max & 13.6                          & 18.9           \\ \cline{2-4} 
\multicolumn{1}{|c|}{}                                                                                                          & avg & 12.9                          & 18.1           \\ \hline
\end{tabular}
\caption{Comparison results of \name and MOSAT}
\label{comparison}
\end{table}

In each 24-hour running, MOSAT can find 6 types of safety violations of Apollo. On average, MOSAT generates 3541 solutions (min 3208 and max 3719), and 57 (min 49 and max 61) out of them detect safety violations of Apollo. 

\name detects 10 distinct types of safety violations of Apollo and all of them are detected in the first 14 hours. MOSAT detects 6 types of safety violations of Apollo and all of them are detected in the first 19 hours. The average Euclidean Distance across the detected different safety violation types of \name is 81.10 meters, and that of $\mu MOEA_r$ is 64.70 meters. Moreover, the 6 types of safety violations are all revealed in the 10 types of safety violation detected by \name. It takes \name less time to detect the first safety violation of Apollo than MOSAT. Therefore, compared with MOSAT, \name can detect more types of safety violations of Apollo in a shorter time. On average, one safety violation of Apollo occurs in 12 solutions generated by \name. MOSAT generates 62 solutions to find one safety violation of Apollo. The safety-violation exposure frequency in \name is higher, which shows that \name can efficiently detect more safety violations of Apollo. 
The comparison results show that \name can more effectively detect more types of safety violations in the MCDL system. 

\section{Related Work}
\subsection{Large Language Models for Reasoning}
Recent advancements in large language models (LLMs) have demonstrated their potential in a variety of tasks \cite{ji2023jm3d}, including quantitative reasoning.
LLMs have tremendous capabilities in solving complex tasks, from quantitative reasoning to understanding natural language \cite{romera2024mathematical}. Early models such as GPT-3 have been shown to generate coherent text and perform well in tasks that require contextual understanding, but they often struggle with more complex reasoning tasks, particularly those that involve multi-step logic or abstract thinking.

To address these limitations, researchers have proposed various approaches to enhance the quantitative reasoning capabilities of LLMs. One notable method is the integration of external knowledge bases and joint multi-cues \cite{mann2020language, ji2023jm3d}, which has been shown to improve the accuracy and depth of reasoning by providing models with additional contextual information. Another approach is the use of prompt engineering \cite{reynolds2021prompt}, where carefully designed prompts guide the model towards better reasoning outcomes. Furthermore, there has been growing interest in the application of LLMs to quantitative reasoning tasks in specific domains, such as mathematical problem-solving, where domain-specific training data can significantly enhance the model performance.

\subsection{Multi-Objective Evolutionary Algorithms}
Multi-objective evolutionary algorithms (MOEAs) have become a prominent tool for solving multi-objective optimization problems due to their ability to find a set of Pareto-optimal solutions in a single run \cite{fonseca1993genetic, ishibuchi2008evolutionary, zitzler2001spea2}. Populations in MOEAs generally evolve through high-performing candidate solutions being mutated or recombined to form the next generation.

One of the most typical MOEAs is the Non-dominated Sorting Genetic Algorithm II (NSGA-II) \cite{zitzler1999multiobjective, deb2002fast}, which introduced key innovations such as fast non-dominated sorting and crowding distance mechanisms to find the optimal solutions and maintain solution diversity. NSGA-II has since become a benchmark for comparing other MOEAs due to its balance between computational efficiency and solution quality, which has led to a better understanding of the trade-offs involved in using MOEAs for different types of multi-objective problems


\section{Conclusion and Discussion}
In this paper, we propose \name, an LLM-empowered adaptive evolutionary search method for MCDL systems. Different from existing MOEAs that detect safety violations of MCDL systems starting by randomly initialized population, \name leverages LLM’s ability in language understanding and quantitative reasoning to better comprehend the evolutionary task and create high-quality solutions for the initial population. Based on these, \name adopts an adaptive multi-objective evolutionary algorithm to efficiently search for optimal and diverse solutions. To navigate the search away from the local optima, when the evolutionary process gets stuck, \name feedbacks the characteristics of iterations into the LLM to facilitate the learning of evolutionary experience and population characteristics. Then it promotes the LLM to generating differential seed solutions for the next generation. We uses \name to detect safety violations of a representative MCDL system, industrial autonomous driving system. Furthermore, we evaluate the performance of \name by ablation experiments and compare it with the SOTA method that uses multi-objective genetic algorithm to search solutions for diverse safety violations of the system. The experimental results show that \name can effectively and efficiently detect safety violations of MCDL systems and surpass the SOTA method.

To leverage the LLM's capability, \name inputs the prompt into GPT-4 by sending the API request, which brings extra time cost for waiting the model output. Moreover, the LLM has a limit on the number of input characters, which limits the potential capability of \name due to the limited number of samples for context learning and thought chain. Currently, \name generates the feedback prompt incrementally and updates the early iterations with latest iterations when the characters of the prompt exceeds the limit.

As future work, we aim to employ a local LLM to reduce the time cost for request, and intelligently select the input examples for the feedback, with the potential to further improve the performance and ability of \name. However, there are two main economic implications for deploying $\mu$MOEA in the real world: (1) operational costs of remote LLM access; and (2) infrastructure costs of local LLM deployment. For MCDL systems with infrequent updates or without complicated test requirements, remote access to GPT may be cost-effective. For defect detection tasks, local training and deployment of an LLM on the target system is generally more economical. 

\section{Acknowledgments}
This work is supported by National Natural Science Foundation of China U20A6003 and Major Project of Institute of Software Chinese Academy of Sciences (ISCAS-ZD-202302). This study is also supported under the RIE2020 Industry Alignment Fund – Industry Collaboration Projects (IAF-ICP) Funding Initiative, as well as cash and in-kind contribution from the industry partner(s). The authors are grateful for the financial support provided by the China Scholarship Council program (Grant No. 202304910498).

\bibliography{aaai25}

\end{document}